\documentclass[letterpaper, 10pt, conference]{ieeeconf}  
\IEEEoverridecommandlockouts                              

\usepackage{booktabs}
\usepackage{graphicx}
\usepackage{array}
\usepackage{cite}
\usepackage{graphicx}
\usepackage{amsmath}
\usepackage{amssymb}
\usepackage{algorithm}
\usepackage[noend]{algorithmic}
\usepackage{flushend}
\usepackage[T1]{fontenc}
\usepackage{aecompl}
\usepackage{xurl}
\usepackage{hyperref}
\usepackage{ifthen}
\usepackage{caption}
\usepackage{multirow}
\usepackage{subcaption}

\usepackage{xcolor}
\definecolor{green}{RGB}{11,155,13}

\title{\bf APPLV: Adaptive Planner Parameter Learning from Vision-Language-Action Model

\author{Yuanjie Lu$^1$, Beichen Wang$^1$, Zhengqi Wu$^2$, Yang Li$^3$, Xiaomin Lin$^2$, Chengzhi Mao$^3$, and Xuesu Xiao$^1$}

\thanks{$^1$Department of Computer Science, George Mason University, Virginia, USA}%
\thanks{$^2$Department of Engineering Science, University of South Florida, Florida, USA}%
\thanks{$^3$Department of Computer Science, Rutgers University, New Jersey, USA }
}

\begin{document}

\maketitle

\begin{abstract} 

Autonomous navigation in highly constrained environments remains challenging for mobile robots. Classical navigation approaches offer safety assurances but require environment-specific parameter tuning; end-to-end learning bypasses parameter tuning but struggles with precise control in constrained spaces. To this end, recent robot learning approaches automate parameter tuning while retaining classical systems' safety, yet still face challenges in generalizing to unseen environments. Recently, Vision-Language-Action (VLA) models have shown promise by leveraging foundation models' scene understanding capabilities, but still struggle with precise control and inference latency in navigation tasks. In this paper, we propose Adaptive Planner Parameter Learning from Vision-Language-Action Model (\textsc{applv}). Unlike traditional VLA models that directly output actions, \textsc{applv} leverages pre-trained vision-language models with a regression head to predict planner parameters that configure classical planners. We develop two training strategies: supervised learning fine-tuning from collected navigation trajectories and reinforcement learning fine-tuning to further optimize navigation performance. We evaluate \textsc{applv} across multiple motion planners on the simulated Benchmark Autonomous Robot Navigation (BARN) dataset and in physical robot experiments. Results demonstrate that \textsc{applv} outperforms existing methods in both navigation performance and generalization to unseen environments.



\end{abstract}


\section{Introduction}
\label{sec:Intro}

Autonomous mobile robot navigation in complex, highly constrained environments remains a fundamental challenge in robotics, with critical applications spanning warehouse logistics, autonomous delivery, and service robotics. The central problem is enabling robots to safely and efficiently navigate narrow passages and cluttered spaces with extremely low clearance while maintaining robust performance across diverse, previously unseen scenarios.

Traditional navigation systems have established themselves as the industry standard due to their reliability and safety assurances. However, they suffer from a critical weakness: performance heavily depends on parameter configurations that must be precisely tuned for each deployment scenario, including velocity limits, cost weights, sampling densities, and safety margins. This creates practical barriers: it requires expertise most end users lack, is time-consuming and environment-specific, and static parameters cannot adapt to varying conditions within a single environment.
To address parameter sensitivity, end-to-end learning methods directly map sensory inputs to control commands, bypassing parameter tuning entirely. While conceptually elegant, these approaches sacrifice the interpretability and safety assurances of classical systems and struggle to generalize beyond the training distribution. In highly constrained environments where tight clearances demand precision, real-world sensor noise and actuation uncertainties make reliable generalization even more challenging.

\begin{figure}[!t]
\centering

\includegraphics[width=0.99\columnwidth]{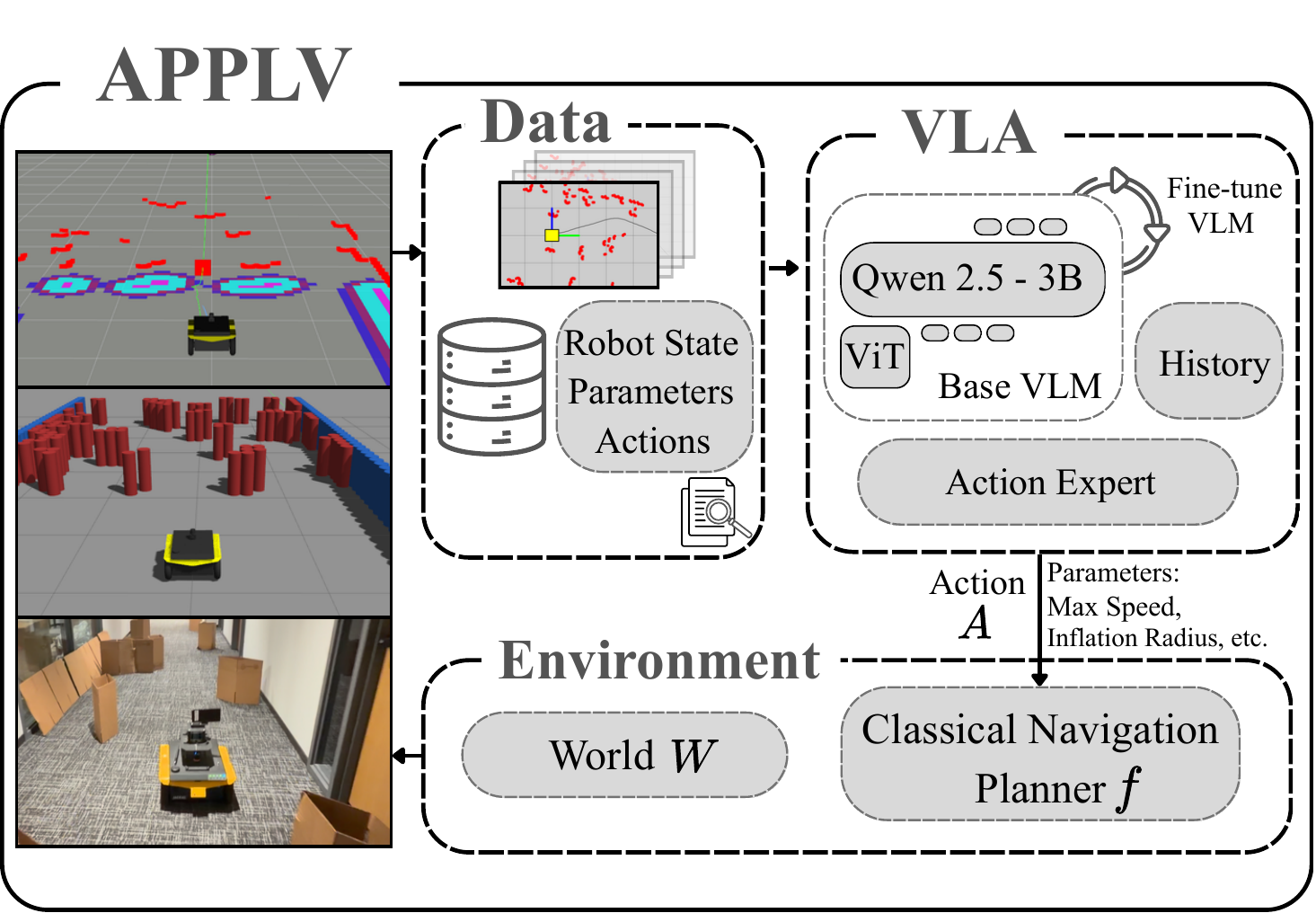}
\caption{\textsc{applv} instantiates the \textsc{appl} paradigm with a VLA model to dynamically adjust a classical navigation planner's  parameters, learned from data collected in simulation.}
\vspace{-3mm}
\label{fig:framework}
\end{figure}


Recognizing the limitations of both extremes, hybrid approaches in robot learning have emerged to combine the safety of classical navigation with learning-based adaptability. The Adaptive Planner Parameter Learning (\textsc{appl}) family~\cite{xiao2022appl} 
exemplifies this paradigm by automating online parameter selection, demonstrating improvements over both manual tuning and end-to-end learning. However, these methods struggle to generalize to unseen environments, with performance degrading even in qualitatively similar scenarios. Moreover, requiring human interactions~\cite{xiao2020appld, wang2021appli, wang2021apple} or random exploration~\cite{xu2021applr}, their overall navigation performance still leaves considerable room for improvement. In challenging spaces like narrow corridors or dense clutter, they often exhibit suboptimal behavior, either showing excessive caution that causes slow traversal or making aggressive choices that compromise safety. 

Recently, vision-language foundation models have demonstrated remarkable capabilities in understanding complex visual scenes and reasoning about spatial relationships. Building on these advances, Vision-Language-Action (VLA) models have emerged as a promising paradigm for robot learning, leveraging pre-trained Vision-Language Models (VLMs) to directly map visual observations to robot actions. However, VLA models face challenges when applied to constrained navigation: they struggle with the centimeter-level precision required in tight spaces and suffer from high inference latency unsuitable for real-time control.

Motivated by these limitations, we propose Adaptive Planner Parameter Learning from Vision-Language-Action Model (\textsc{applv}, Fig.~\ref{fig:framework}), which leverages a vision-language-action model to predict navigation planner parameters rather than actions directly. Specifically, a VLM processes custom observations to extract multi-layer hidden states, while a history encoder captures temporal context from prior frames. These representations are fed into an action expert comprising a Dense Prediction Transformer (\textsc{DPT})-style regression head~\cite{ranftl2021vision} that fuses multi-layer features and temporal context to predict parameters such as velocity limits, sampling densities, and cost weights. These parameters configure a classical navigation planner to generate executable robot actions, enabling fine-grained environmental adaptation while maintaining safety assurances and computational efficiency. We propose two training strategies for \textsc{applv}: supervised learning fine-tuning from collected navigation trajectories (\textsc{applv-sl}) and reinforcement learning fine-tuning (\textsc{applv-rlft}) to further optimize navigation performance. We evaluate \textsc{applv} across four local planners on the simulated Benchmark Autonomous Robot Navigation (BARN) dataset~\cite{perille2020benchmarking, xiaoautonomous} and in physical robot experiments, demonstrating superior performance over existing parameter learning methods in both navigation performance and generalization to unseen environments.


\section{Related Work}

We review related work on autonomous robot navigation, tracing the evolution from classical parameter tuning to end-to-end learning methods, and more recently to hybrid approaches and foundation models.

Classical navigation systems require manual parameter tuning for each deployment~\cite{guimaraes2016ros, das2024motion, lu2025multi,zheng2021ros}. While automated optimization approaches using fuzzy logic~\cite{castillo2015new} and neuro-fuzzy systems~\cite{teso2019predictive} reduce manual effort, they remain fundamentally limited 
due to their assumption that a single set of optimal parameters should fit everywhere in a deployment environment. 
This has led researchers to explore end-to-end learning methods that bypass parameter tuning entirely.

To bypass parameter tuning, end-to-end navigation methods directly map sensory inputs to control commands. Imitation learning approaches~\cite{ codevilla2018end} train navigation policies from expert demonstrations, and reinforcement learning methods~\cite{xu2025verti,faust2018prm} learn through trial-and-error interaction with environments. However, these approaches sacrifice the interpretability and safety assurances of classical systems. Moreover, they struggle to generalize beyond training distributions, particularly in highly constrained environments where tight clearances demand centimeter-level precision and real-world sensor noise creates additional challenges.

To address the limitations of both parameter tuning and end-to-end learning, hybrid approaches combine classical navigation methods with learning-based parameter selection. Neurofuzzy methods~\cite{zhu2007neurofuzzy} use neural networks to tune controller parameters. Reinforcement learning approaches adapt cost functions~\cite{wang2025reward} or dynamics parameters~\cite{lu2025adaptive, lu2023leveraging, bui2022improving} to varying environmental complexity. The \textsc{appl} family~\cite{xiao2022appl} enables dynamic parameter selection during navigation. \textsc{appld}~\cite{xiao2020appld} learns parameter sets from demonstrations, \textsc{appli}~\cite{wang2021appli} from corrective interventions, \textsc{apple}~\cite{wang2021apple} from evaluative feedback, and \textsc{applr}~\cite{xu2021applr} uses reinforcement learning to learn a parameter policy. Although these approaches improve upon both manual parameter tuning and end-to-end learning, significant challenges remain: they struggle to generalize to unseen environments, and overall navigation performance in highly constrained spaces still leaves substantial room for improvement~\cite{xiao2023autonomous, xiao2024autonomous, xiaoautonomous, xiao2026autonomous}.

Recently, VLMs trained on large-scale image-text data have demonstrated strong visual understanding capabilities. Building on this success, VLA models~\cite{kim2024openvla, black2024pi_0} adapt these foundation models to robotics by fine-tuning them to directly predict robot actions. For navigation, approaches such as LM-Nav~\cite{shah2023lm}, NaVILA~\cite{cheng2024navila}, and OmniVLA~\cite{hirose2025omnivla} apply VLMs to instruction following and multi-modal goal specifications. While these foundation model approaches demonstrate strong language grounding and scene understanding, their use in constrained navigation remains unexplored. Designed for open spaces with high computational tolerance, they fail to deliver the centimeter-level precision demanded by narrow passages and cluttered environments, and their large model sizes incur inference latency incompatible with real-time control.

Our approach, \textsc{applv}, explores an alternative direction: instead of directly predicting actions, we leverage vision-language models to predict low-level navigation parameters for classical planners, enabling navigation in highly constrained spaces. By operating at the parameter level, \textsc{applv} provides fine-grained adaptation to environmental constraints while preserving the safety assurances and interpretability of classical systems. Unlike direct action prediction that demands high-frequency inference at every control step, parameter prediction operates at a much lower frequency since classical planners can autonomously produce real-time actions under the same set of parameters, significantly relaxing real-time computational constraints.
\section{Preliminaries}
\label{sec:Preliminaries}
\textsc{applv} leverages pre-trained VLMs with superior reasoning capabilities further fine-tuned on robot navigation data within the \textsc{appl} paradigm to enhance the \textsc{appl} parameter policy's ability to dynamically adjust planners' parameters. 

\subsection{The \textsc{appl} Paradigm}
The \textsc{appl} paradigm assumes a mobile robot equipped with a classical navigation planner $f: \mathcal{X} \times \Phi \rightarrow \mathcal{A}$, where $\mathcal{X}$ is the planner's state space (e.g., robot odometry, sensory inputs, and navigation goal), $\Phi$ is the parameter space (e.g., velocity limits, cost coefficients, and inflation radius), and $\mathcal{A}$ is the action space (e.g., linear velocity $v$ and angular velocity $\omega$). Given a state $x \in \mathcal{X}$ and planner parameters $\phi \in \Phi \subset \mathbb{R}^d$ with $d$ depending on the target planner, the planner generates an action $a = f(x; \phi)=f_\phi(x)$. \textsc{appl} formulates parameter learning in a meta-MDP with a meta-environment $\mathcal{E}$ that combines the physical world $\mathcal{W}$ and planner $f$. The meta-state $s_t = (x_t, \phi_{t-1}) \in \mathcal{S} = \mathcal{X} \times \Phi$ includes both the planner's state and previous parameters, while the meta-action is the parameter configuration $\phi_t \in \Phi$. An \textsc{appl} agent learns a parameter policy $\pi_p: \mathcal{S} \rightarrow \Phi$ that selects parameters to maximize expected cumulative reward, instead of the conventional motion policy $\pi_m: \mathcal{X} \rightarrow \mathcal{A}$ that issues raw motor commands. Building upon the classical navigation planner $f$, \textsc{appl} inherits $f$'s safety and explainability.

\subsection{Vision-Language Models}
VLMs learn joint representations of visual and linguistic information by pre-training on large-scale image-text datasets. A typical VLM consists of a visual encoder, a projection module, and a Large Language Model (LLM) backbone, trained end-to-end with a next-token prediction objective. Qwen2.5-VL~\cite{bai2025qwen25vltechnicalreport} follows this 
architecture, encoding image inputs as patch embeddings that are projected and concatenated with text tokens before being processed by the LLM. VLMs can be efficiently adapted to downstream tasks via Low-Rank Adaptation (LoRA)~\cite{hu2021loralowrankadaptationlarge}, which inserts trainable low-rank matrices into the transformer layers while keeping pre-trained weights frozen.

\section{APPLV}

Fig.~\ref{fig:applv_framework} illustrates the \textsc{applv} architecture. The meta-state is $s_t = (x_t, \phi_{t-1})$, where the planner's state $x_t$ comprises a current custom image $I_t$, historical images from previous frames $\{I_{t-k}\}_{k=1}^{K}$, and current robot state including linear and angular velocities, while $\phi_{t-1}$ are the previous planner parameters. A Qwen2.5-VL-3B model processes the current image with a language prompt, extracting multi-layer hidden states. These features are concatenated with features from the history encoder and fed into a DPT regression head to predict planner parameters $\phi_t$, which configure the classical navigation planner $f$ to generate motion control commands.

\subsection{The \textsc{applv} Model Architecture}
\subsubsection{Training Dataset}

\begin{figure}[!t]
  \centering  
  \includegraphics[width=0.99\columnwidth]{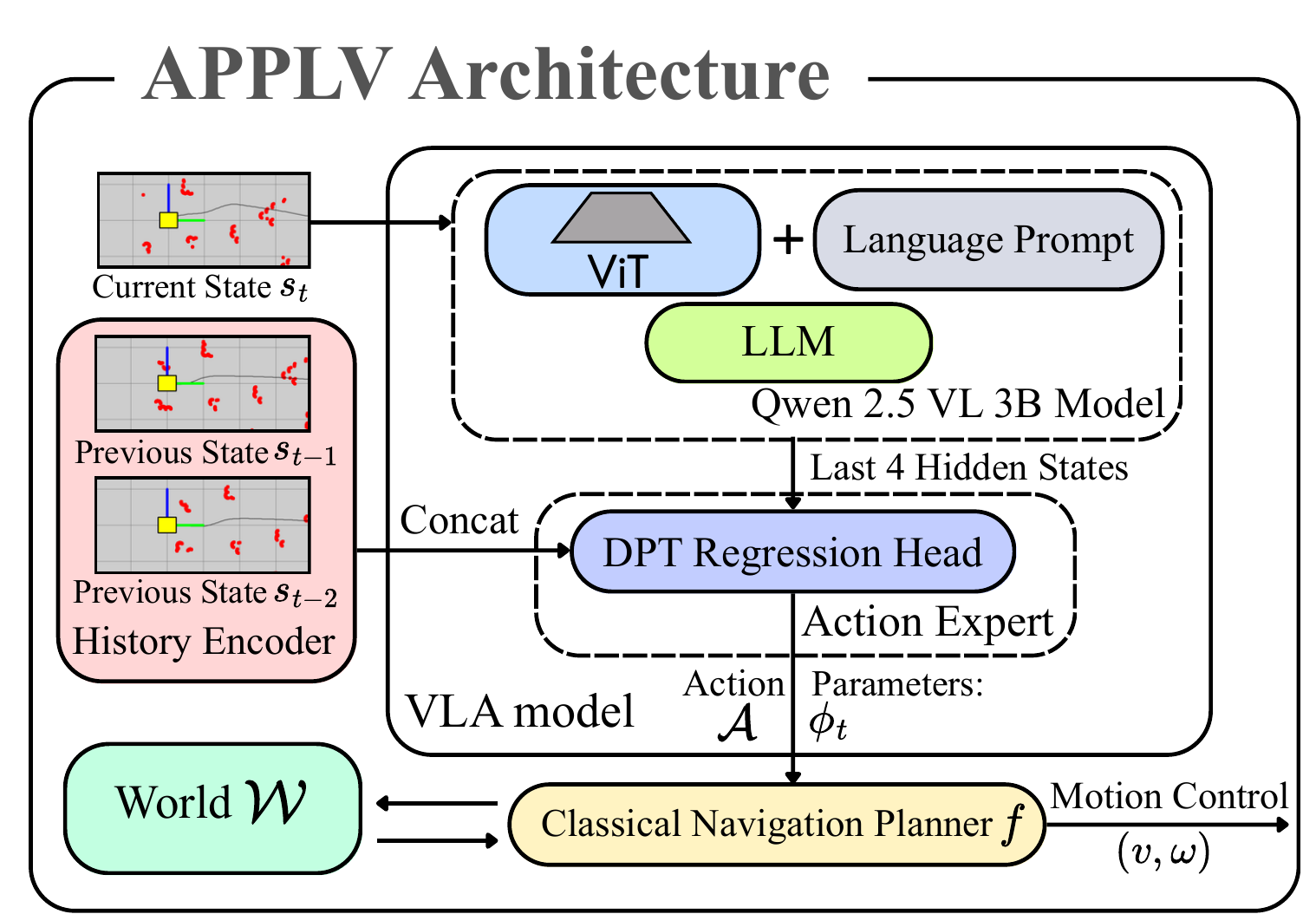}\\[-2mm]
  \caption{\textsc{applv} Architecture. A VLA model processes current and historical states to predict planner parameters, which configure a classical navigation planner to generate motion control commands.}
  \vspace{-7mm}
  \label{fig:applv_framework}
\end{figure}

Each training sample consists of a custom image $I_t$, historical frames $\{I_{t-k}\}_{k=1}^{K}$ for temporal context, robot current linear and angular velocity state encoded in the text prompt, and corresponding planner parameters $\phi_t$. The custom image (Fig.~\ref{fig:applv_framework}, top left) is a top-down RGB representation of the robot's local environment with:
\begin{itemize}
  \item \textbf{Gray background}: environment background in gray;
  \item \textbf{Red overlay}: laser scan of obstacles;
  \item \textbf{Blue path}: global path from the robot to its goal; and
  \item \textbf{Robot marker}: robot footprint in the base link frame.
\end{itemize}
Samples are stored in JSON format and shuffled during training. The demonstration dataset for supervised learning comprises data from two sources: (1) hand-crafted heuristic rules designed by an expert roboticist that map perceived environmental conditions (e.g., obstacle density and corridor width) to predefined parameter configurations, and (2) policies learned by the \textsc{applr}~\cite{xu2021applr} baseline method. Both data collection approaches are detailed in Section~\ref{sec:Experiment}.

\subsubsection{Vision-Language Backbone}
Qwen2.5-VL-3B is employed to extract visual-semantic representations from custom images and text prompts. The model consists of a Vision Transformer (ViT) encoder~\cite{dosovitskiy2021imageworth16x16words} and a language model backbone. Unlike standard VLM applications that generate text outputs, intermediate hidden states are extracted for parameter regression. Hidden states from the last four transformer layers are extracted as $\{\mathbf{h}^{(l)}_t\}_{l=L-3}^{L}$, where each $\mathbf{h}^{(l)}_t \in \mathbb{R}^{N \times 256}$ represents $N$ tokens. These multi-layer features capture spatial patterns at different levels of abstraction. During training, the vision encoder is frozen, while the language model is adapted using LoRA with rank $r=64$ and scaling factor $\alpha=128$ to enable parameter-efficient fine-tuning.

\begin{figure*}[!t]
  \centering  
  \includegraphics[width=0.95\textwidth]{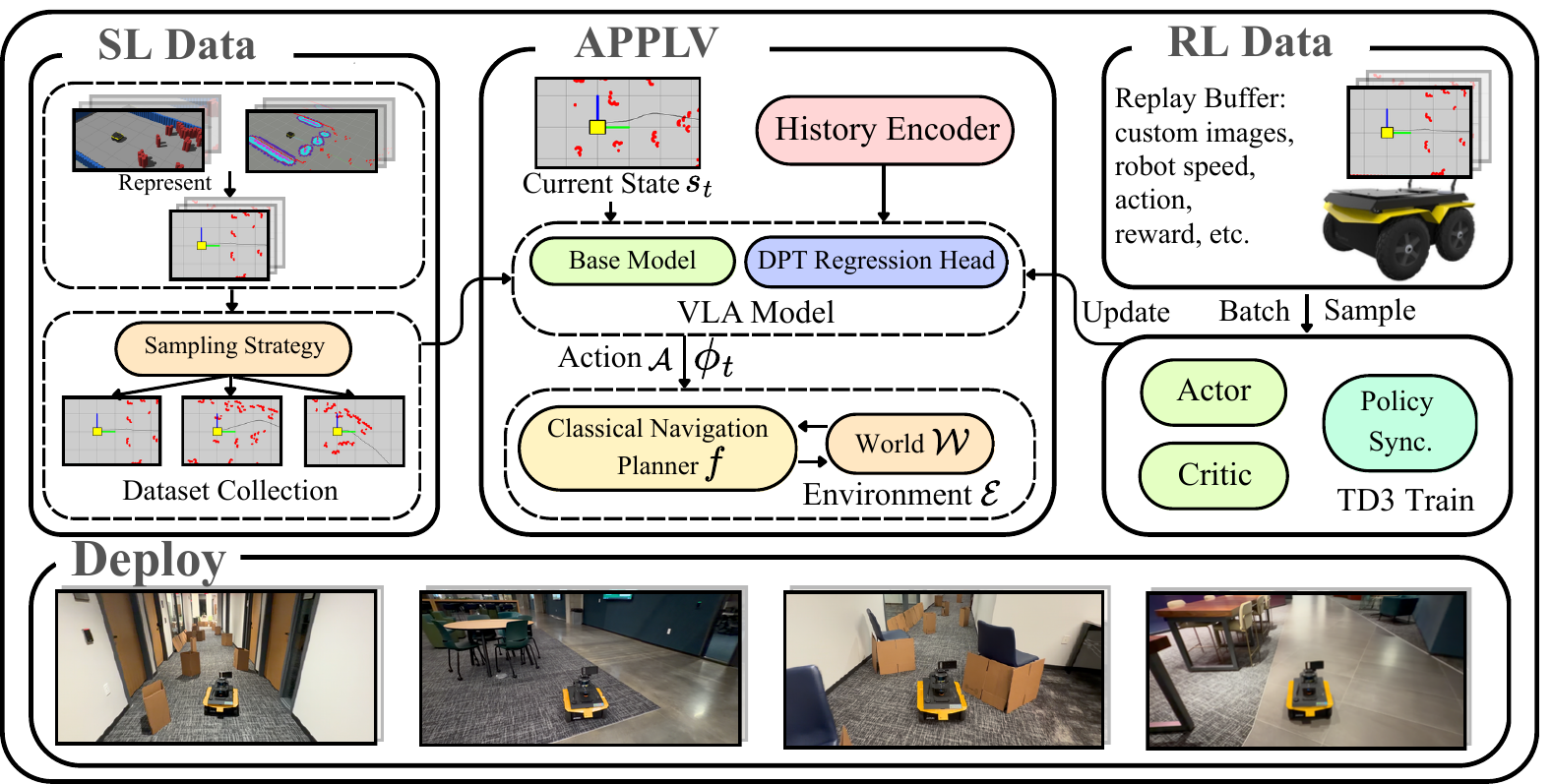}\\[-1mm]
  \caption{\textsc{applv} Training and Deployment Pipeline. Left: Supervised learning data collection with representation and
  sampling. Middle: \textsc{applv} architecture with VLA model and classical planner. Right: Reinforcement Learning fine-tuning with TD3. Bottom: Physical deployment in natural cluttered environments.}
  \vspace{-6mm}
  \label{fig:framework_2}
\end{figure*}

\subsubsection{History Encoder}
To incorporate temporal context, a lightweight history encoder processes prior visual frames. A convolutional network extracts spatial features from each frame independently. The frame-level features are augmented with learnable positional embeddings and fed into a temporal transformer encoder that models sequential dependencies across frames. The encoded historical representation is fused with the current frame's VLM features in the DPT head, enabling temporally-aware parameter predictions.

\subsubsection{DPT Regression Head}
The DPT regression head~\cite{ranftl2021vision} serves as the action expert, fusing multi-layer hidden states with encoded historical features to predict planner parameters. The architecture reassembles tokens from different transformer stages into feature representations and progressively combines them through convolutional fusion. Each layer's hidden states $\mathbf{h}^{(l)}_t$ are projected to a unified feature space: $\mathbf{f}^{(l)}_t \in \mathbb{R}^{N \times 256}$. Features are fused progressively from deeper to shallower layers using residual convolutional blocks, with the historical context concatenated during fusion. The fused representation undergoes attention-weighted pooling to produce a global feature vector $\mathbf{z}_t \in \mathbb{R}^{256}$. A final MLP regresses the parameter vector: $\phi_t = \text{MLP}_{\text{reg}}(\mathbf{z}_t)$, where $\phi_t \in \mathbb{R}^d$ with $d$ depending on the target planner.

\subsubsection{Classical Navigation Planner}
The classical navigation planner receives predicted parameters $\phi_t$ and generates motion control commands $(v_t, \omega_t)$. Different planner implementations share a common parameterized structure where $\phi_t$ configures behavior-critical aspects such as velocity limits, cost function weights, planning horizons, and inflation radius. Based on these parameters, the planner selects optimal actions while ensuring kinodynamic feasibility and collision avoidance. This design allows the VLA model to adapt planning behavior across diverse scenarios and planner types through learned parameter selection.

\subsection{Training Strategies}
\subsubsection{Supervised Learning (\textsc{applv-sl})}
The base VLA model is trained via Behavior Cloning (BC) on demonstration trajectories (Fig.~\ref{fig:framework_2} left). As illustrated, raw demonstration data is first processed through a representation step that converts environmental observations into custom images. A sampling strategy then selects diverse training samples to construct the final dataset. Given a dataset $\mathcal{D}$ of such samples, the training objective minimizes mean squared error:
\begin{equation}
    \mathcal{L}_{\text{SL}} = \frac{1}{|\mathcal{D}|} \sum_{i=1}^{|\mathcal{D}|} \|\phi_i - \phi_i^*\|^2_2, \nonumber
\end{equation}
where $\phi_i$ is the predicted parameter for sample $i$ and $\phi_i^*$ is the corresponding ground-truth parameter from demonstration data.

\subsubsection{Reinforcement Learning Fine-Tuning (\textsc{applv-rlft})}
The second training strategy employs off-policy reinforcement learning (Fig.~\ref{fig:framework_2} right). The VLA model can be initialized from the supervised pretrained weights or trained from scratch; we adopt the former for accelerated convergence. The input and output spaces remain identical to supervised learning. A distributed collection framework deploys multiple parallel collectors, each running the latest policy to interact with an independent simulation environment. Experiences are aggregated into a shared replay buffer. The reward function encourages effective navigation through four components: (1) progress reward $r_{\text{progress}}$ measuring distance reduction toward the goal, (2) collision penalty $r_{\text{collision}}$ enforcing safety constraints, (3) time penalty $r_{\text{time}}$ promoting efficient task completion, and (4) obstacle avoidance reward $r_{\text{obstacle}}$ encouraging safe proximity to environmental obstacles. The total reward is:
$
r_t = r_{\text{progress}} + r_{\text{collision}} + r_{\text{time}} + r_{\text{obstacle}}.
$ 
The VLA model serves as the actor, predicting planner parameters from visual and state inputs. A separate critic network estimates the value of the predicted parameters. Training follows the Twin Delayed Deep Deterministic Policy Gradient (TD3) algorithm~\cite{fujimoto2018addressing}: the critic is updated every step by minimizing temporal difference error between predicted and target Q-values, while the actor is updated every two steps by maximizing the critic's Q-value estimate. Periodic policy synchronization ensures collectors use recent policies, balancing exploration and exploitation.

\section{Experiments}
\label{sec:ExpResults}

The BARN Challenge benchmark focuses on highly constrained scenarios with narrow passages and dense obstacles. We integrate \textsc{applv} with four representative local planners: DWA~\cite{fox2002dynamic}, TEB~\cite{rosmann2012trajectory}, MPPI~\cite{williams2017model}, and DDP~\cite{lu2025decremental}. \textsc{applv} is compared against baselines spanning heuristics, BC, reinforcement learning, and zero-shot foundation model approaches. Experiments are conducted across 300 simulated BARN environments~\cite{perille2020benchmarking}, with additional real-world validation on a Clearpath Jackal robot.

\subsection{Experimental Setup}
\label{sec:Experiment}
\subsubsection{Navigation Planners}
To demonstrate that \textsc{applv} generalizes across different navigation planners, we integrate it with four representative local planners: DWA~\cite{fox2002dynamic}, which samples velocity commands within the robot's dynamic window; TEB~\cite{rosmann2012trajectory}, which optimizes an elastic band of timed poses under kinodynamic constraints; MPPI~\cite{williams2017model}, a sampling-based planner that computes optimal controls via weighted averaging of stochastic rollouts; and DDP~\cite{lu2025decremental}, a recently proposed dynamics-based planner that achieved second place in the 2025 BARN Challenge~\cite{xiaoautonomous}. DWA and TEB are integrated from the ROS move\_base stack, while MPPI and DDP are custom C++ implementations.

\begin{figure}[!t]
\centering
\includegraphics[width=0.95\columnwidth]{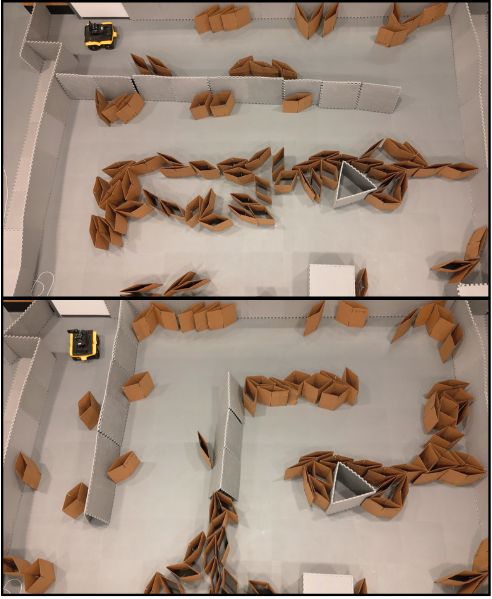}
\caption{Physical deployment of two BARN Challenge environments}
\vspace{-3mm}
\label{fig:physical}
\end{figure}

\subsubsection{Baseline Methods}
We compare \textsc{applv} against four baseline methods for navigation parameter prediction:

\textbf{Heuristic Expert} represents the traditional approach to parameter configuration in classical navigation systems. An expert roboticist designs hand-crafted rules that map perceived environmental conditions, such as obstacle density and corridor width, to predefined parameter configurations. This baseline reflects common practice in real-world deployments and establishes a practical comparison for learning-based methods.

\begin{table*}
    \centering
    \caption{Performance Comparison of Methods Grouped by Planners on 300 Test BARN Environments.}
    \setlength{\tabcolsep}{12pt}
    \begin{tabular}{llccccc}
        \toprule
        \textbf{Planner} & \textbf{Method} & \textbf{Success (\%)} $\uparrow$ & \textbf{Avg. Time (s)} $\downarrow$ & \textbf{Avg. Score} $\uparrow$ & \textbf{Collision (\%)} $\downarrow$ & \textbf{Timeout (\%)} $\downarrow$ \\
        \midrule
        \multirow{6}{*}{DWA}
        & \textsc{applr}            & 73.15          & 27.38          & 0.296          & 11.87          & 14.98 \\
        & Heuristic Expert & 82.47          & 25.83          & 0.328          & 07.50          & 10.03 \\
        & Transformer BC   & 83.03          & 27.58          & 0.323          & 08.21          & 08.75 \\
        & Zero-Shot VLM    & 81.00          & 31.27          & 0.301          & 10.00          & 09.00 \\
        & \textsc{applv-sl}             & 86.10          & 21.58          & 0.357          & 07.51          & 06.39 \\
        & \textsc{applv-rlft}       & \textbf{87.20} & \textbf{18.68} & \textbf{0.374} & \textbf{06.51} & \textbf{06.29} \\
        \midrule
        \multirow{6}{*}{MPPI}
        & \textsc{applr}            & 78.53          & 24.73          & 0.356          & 07.40          & 14.07 \\
        & Heuristic Expert & 84.48          & 23.41          & 0.365          & 14.17          & \textbf{01.35} \\
        & Transformer BC   & 83.68          & 20.42          & 0.378          & 06.52          & 09.80 \\
        & Zero-Shot VLM    & 85.24          & 22.02          & 0.367          & 08.02          & 06.74 \\
        & \textsc{applv-sl}         & 88.93          & 17.82          & 0.415          & 09.84          & 01.22 \\
        & \textsc{applv-rlft}       & \textbf{89.70} & \textbf{16.75} & \textbf{0.434} & \textbf{07.30} & 03.00 \\
        \midrule
        \multirow{6}{*}{TEB}
        & \textsc{applr}            & 79.17          & 20.29          & 0.362          & 05.53          & 15.30 \\
        & Heuristic Expert & 86.61          & 19.11          & 0.388          & \textbf{00.58} & 12.81 \\
        & Transformer BC   & 90.25          & 20.35          & 0.383          & 02.10          & 07.64 \\
        & Zero-Shot VLM    & 89.31          & 16.64          & 0.398          & 01.07          & 09.61 \\
        & \textsc{applv-sl}         & 90.00          & 17.61          & 0.429          & 00.78          & 09.22 \\
        & \textsc{applv-rlft}       & \textbf{90.30} & \textbf{12.51} & \textbf{0.441} & 08.90          & \textbf{00.70} \\
        \midrule
        \multirow{6}{*}{DDP}
        & \textsc{applr}            & 85.35          & 15.66          & 0.404          & 07.93          & 06.72 \\
        & Heuristic Expert & 89.50          & 16.09          & 0.418          & 02.38          & 08.12 \\
        & Transformer BC   & 85.57          & 18.16          & 0.411          & 02.36          & 12.05 \\
        & Zero-Shot VLM    & 92.50          & 16.18          & 0.417          & \textbf{00.50} & 04.00 \\
        & \textsc{applv-sl}         & 92.68          & 15.28          & 0.440          & 01.66          & 05.65 \\
        & \textsc{applv-rlft}       & \textbf{94.34} & \textbf{13.63} & \textbf{0.463} & 01.98          & \textbf{03.68} \\
        \bottomrule
    \end{tabular}
    \label{tab:comparison_by_planner}

    \vspace{4pt}
    \parbox{\textwidth}{\raggedright Each environment is tested 3 times. Bold indicates the best result within each planner group.}
    \vspace{-5mm}
\end{table*}

\textbf{\textsc{applr}}~\cite{xu2021applr}, a representative method from the APPL family~\cite{xiao2022appl}, trains a TD3-based reinforcement learning policy that maps laser scan observations to navigation parameters. This purely RL-based approach has demonstrated substantial improvements over static parameter configurations in classical planners like DWA, serving as a strong baseline for learned parameter adaptation.

\textbf{Transformer BC} serves as a controlled baseline to isolate the contribution of vision-language pre-training. We train a transformer model from scratch with a comparable number of parameters and the same training data as \textsc{applv}, using supervised learning on collected demonstrations to predict navigation parameters. Unlike \textsc{applv} which fine-tunes a pre-trained VLM, Transformer BC uses random initialization, allowing us to quantify the performance gain from leveraging vision-language pre-training.

\textbf{Zero-Shot VLM} evaluates whether foundation models can solve navigation parameter prediction without task-specific training. We prompt GPT-4o~\cite{openai2024gpt4o} with custom images and robot state to directly generate navigation parameters as text output, relying entirely on the model's zero-shot spatial reasoning and numerical estimation capabilities. This baseline demonstrates the performance gap between zero-shot prompting and task-specific fine-tuning, validating the necessity of adapting foundation models to navigation tasks.

\subsubsection{Experimental Environments}
Our evaluation encompasses both simulated and real-world scenarios. The BARN benchmark~\cite{perille2020benchmarking} provides 300 highly constrained environments with diverse obstacle configurations generated via cellular automata for training (Fig.~\ref{fig:framework_2}, middle left). An additional set of 300 entirely different environments is used for testing to evaluate generalization to unseen scenarios. Physical experiments are
conducted in indoor test courses using a Clearpath Jackal robot (Fig.~\ref{fig:framework_2} bottom; Fig.~\ref{fig:physical}).

\subsubsection{Pre-Train Dataset Collection}
Training data is collected using both Heuristic Expert and \textsc{applr} parameters in simulation environments. For each of the four planners, we execute 1000 trajectory attempts per environment across all 300 BARN training environments, collecting both successful and failed navigation experiences. During each trajectory execution, we record per-step observations at 0.5-second intervals, including custom images, robot velocities, and corresponding planner parameters. Each trajectory contains multiple time steps, forming a sequence of navigation experiences. To ensure dataset quality and balance, we employ two sampling strategies: difficulty-aware sampling and motion-aware filtering. Difficulty-aware sampling classifies environments into four difficulty tiers based on average navigation score from the BARN Challenge. We apply progressively relaxed score thresholds across tiers: easier environments retain only high-scoring trajectories to focus on successful behaviors, while harder environments gradually lower the threshold to admit broader trajectory distributions, ensuring sufficient training data diversity in challenging scenarios where high-scoring trajectories are scarce. Motion-aware filtering evaluates whether each frame makes progress toward the goal by measuring the distance reduction to the target. Frames with forward progress (i.e., distance to goal decreases) are fully retained as they demonstrate effective navigation behavior. In contrast, stagnant or rotational frames (i.e., distance to goal unchanged or increasing) are subsampled at approximately 10\% retention rate, as they provide limited learning value while still capturing obstacle avoidance behaviors. After applying both sampling strategies, we obtain approximately 30,000 high-quality training samples per planner.


\subsubsection{Training Parameters Configuration}
The \textsc{applv} model processes 600×400 pixel custom images, which together with text prompts occupy approximately 601 tokens as input to the VLM. The model consists of three main components: LoRA adapters (330M parameters) applied to the pre-trained vision-language backbone, a DPT-style regression head (3.8M parameters), and a history encoder (1.7M parameters). During supervised pre-training, we train the model with a batch size of 4, optimizing the LoRA adapters, history encoder, and regression head while keeping the VLM backbone frozen. For Reinforcement Learning Fine-Tuning (RLFT), we adopt a TD3-based framework where the actor network is initialized from the supervised learning checkpoint. The critic network has the same architecture and parameter count as the actor, but with independent parameters. Unlike \textsc{applr}~\cite{xu2021applr}, which uses a 2-layer MLP to map 721-dimensional observations (720 laser rays + goal angle) to navigation parameters, our actor processes custom images through the full \textsc{applv} architecture with vision-language representations. We use ZeRO-2 optimization with a batch size of 42 during RLFT.

\subsubsection{Software and Hardware Configuration}
All simulation experiments are conducted on a SLURM cluster using Singularity containers~\cite{kurtzer2017singularity}. The simulation environment runs Ubuntu 20.04 with ROS Noetic and Python 3.8, executing 250-300 parallel Gazebo instances. Simulations communicate with FastAPI inference servers on A100 GPU nodes. Physical experiments use a Clearpath Jackal robot with a Hokuyo LiDAR sensor (720-dimensional scans, 270° field of view). The robot's onboard computer sends observations to an RTX 5070 Ti GPU server via FastAPI for real-time parameter prediction. Model training and inference use Python 3.10. The \textsc{applv} model is built on Qwen2.5-VL-3B. Inference latency per prediction is approximately 0.41s on RTX 5070 Ti, 0.47s on RTX 3080, and 0.27s on RTX 5090.

\subsection{Experimental Results}
We evaluate navigation performance in simulation using the BARN Challenge scoring metric~\cite{xiaoautonomous}, which balances success rate and traversal time efficiency. Physical experiments use success rate, completion progress, and traversal time as metrics.
\subsubsection{Simulated BARN Environments}
Table~\ref{tab:comparison_by_planner} presents the performance of \textsc{applv} against baseline methods across four local planners on 300 unseen BARN test environments. Both \textsc{applv-sl} and \textsc{applv-rlft} demonstrate substantial improvements over all baselines in terms of success rate, navigation time, and overall navigation score. To isolate the contribution of vision-language pre-training, we compare \textsc{applv-sl} against Transformer BC, which uses a comparable model size and the same training data but is trained from scratch. \textsc{applv-sl} consistently outperforms Transformer BC across all planners, validating that leveraging pre-trained VLMs significantly improves performance. Comparing \textsc{applv-sl} with Zero-Shot VLM (GPT-4o prompted to output parameters) evaluates whether fine-tuning is necessary. While Zero-Shot VLM demonstrates reasonable performance, \textsc{applv-sl} achieves superior results, particularly with simpler planners. This indicates that task-specific fine-tuning remains essential despite foundation models' spatial reasoning capabilities. \textsc{applv-rlft} consistently improves upon \textsc{applv-sl} across all planners, demonstrating that RLFT effectively optimizes for task objectives beyond supervised imitation. Compared to the reinforcement-learning-only baseline \textsc{applr}, \textsc{applv-rlft} achieves substantially higher success rates and faster navigation across all planners, validating that vision-language representations significantly outperform laser-scan-based approaches. 

  \begin{table}[t]
    \centering
    \renewcommand{\arraystretch}{1.2}
    \caption{Performance in Test Physical Environments.}
    \setlength{\tabcolsep}{4pt}
    \begin{tabular}{llccc}
        \toprule
        \textbf{Planner} & \textbf{Method} & \textbf{Success} $\uparrow$ & \textbf{Progress} (\%) $\uparrow$ & \textbf{Time (s)}
   $\downarrow$ \\
        \midrule
        \multirow{5}{*}{DWA}
        & \textsc{applr}   & 0/6 & 13.00 $\pm$ 3.2 & 120.00 $\pm$ 0.0 \\
        & Heuristic Expert      & 0/6 & 37.00 $\pm$ 4.5 & 120.00 $\pm$ 0.0 \\
        & Transformer BC     & 0/6 & 40.00 $\pm$ 3.8 & 120.00 $\pm$ 0.0 \\
        & \textsc{applv-sl}  & 0/6 & 64.00 $\pm$ 2.7 & 120.00 $\pm$ 0.0 \\
        & \textsc{applv-rlft}  & \textbf{1/6} & \textbf{75.00 $\pm$ 3.5} & \textbf{95.00 $\pm$ 2.9} \\
        \midrule
        \multirow{5}{*}{MPPI}
        & \textsc{applr}   & 1/6 & 32.00 $\pm$ 4.1 & 102.00 $\pm$ 3.6 \\
        & Heuristic Expert      & 3/6 & 73.00 $\pm$ 2.8 & 80.00 $\pm$ 4.7 \\
        & Transformer BC     & 2/6 & 71.00 $\pm$ 3.4 & 61.00 $\pm$ 2.5 \\
        & \textsc{applv-sl}  &  \textbf{6/6} & \textbf{100.00 $\pm$ 0.0} & 40.00 $\pm$ 3.9 \\
        & \textsc{applv-rlft}  &  \textbf{6/6} & \textbf{100.00 $\pm$ 0.0} & \textbf{34.00 $\pm$ 2.3} \\
        \midrule
        \multirow{5}{*}{TEB}
        & \textsc{applr}   & 0/6 & 18.00 $\pm$ 4.2 & 120.00 $\pm$ 0.0 \\
        & Heuristic Expert      & 1/6 & 32.00 $\pm$ 3.6 & 105.00 $\pm$ 4.8 \\
        & Transformer BC     & 1/6 & 34.00 $\pm$ 2.9 & 103.00 $\pm$ 3.7 \\
        & \textsc{applv-sl}  & \textbf{2/6} & 64.00 $\pm$ 4.4 & 83.00 $\pm$ 2.6 \\
        & \textsc{applv-rlft}  & \textbf{2/6} & \textbf{87.00 $\pm$ 3.1 }& \textbf{67.00 $\pm$ 4.5} \\
        \midrule
        \multirow{5}{*}{DDP}
        & \textsc{applr}   & 0/6 & 22.00 $\pm$ 3.8 & 120.00 $\pm$ 0.0 \\
        & Heuristic Expert      & 2/6 & 78.00 $\pm$ 2.4 & 56.00 $\pm$ 3.3 \\
        & Transformer BC     & 3/6 & 76.00 $\pm$ 4.6 & 64.00 $\pm$ 2.7 \\
        & \textsc{applv-sl}  &  \textbf{6/6} & \textbf{100.00 $\pm$ 0.0} & 34.00 $\pm$ 4.1 \\
        & \textsc{applv-rlft}  &  \textbf{6/6} & \textbf{100.00 $\pm$ 0.0} &  \textbf{32.00 $\pm$ 3.5} \\
        \bottomrule
    \end{tabular}
    \label{tab:real_world}
    \vspace{3pt}
    \parbox{\columnwidth}{\raggedright\footnotesize Each environment is tested 3 times. Bold indicates the best result within each planner group.}
    \vspace{-9.7mm}
  \end{table}

\subsubsection{Physical Experiments}

We conduct physical robot experiments in two distinct real-world environments, with each configuration tested three times. Performance is evaluated by success rate, average progress percentage, and average completion time. Progress percentage measures the distance covered toward the goal, reaching 100\% upon successful arrival. Failed or timed-out trials are assigned a maximum time of 120 seconds. We exclude Zero-Shot VLM baselines from physical experiments due to their prohibitively long inference time. Table~\ref{tab:real_world} presents the physical experiment results across four planners. \textsc{applv} consistently outperforms all three baselines (\textsc{applr}, Heuristic Expert, and Transformer BC) across all planners, with \textsc{applv-rlft} surpassing \textsc{applv-sl} in all cases. Notably, DWA and TEB from ROS move\_base exhibit significantly degraded performance in physical deployment compared to simulation, while custom C++ implementations of MPPI and DDP maintain strong performance. This discrepancy arises from fundamental architectural differences: DWA and TEB rely heavily on costmap-based global path planning, where even minor localization errors can cause severe costmap distortions. When the inflation radius becomes misconfigured due to these errors, the global planner may fail to generate valid paths, causing navigation failure. In contrast, custom MPPI and DDP implementations depend primarily on direct laser scan observations and maintain a history of global paths, making them more robust to transient localization errors in physical environments.

\begin{figure}[!t]
  \centering  
  \includegraphics[width=0.99\columnwidth]{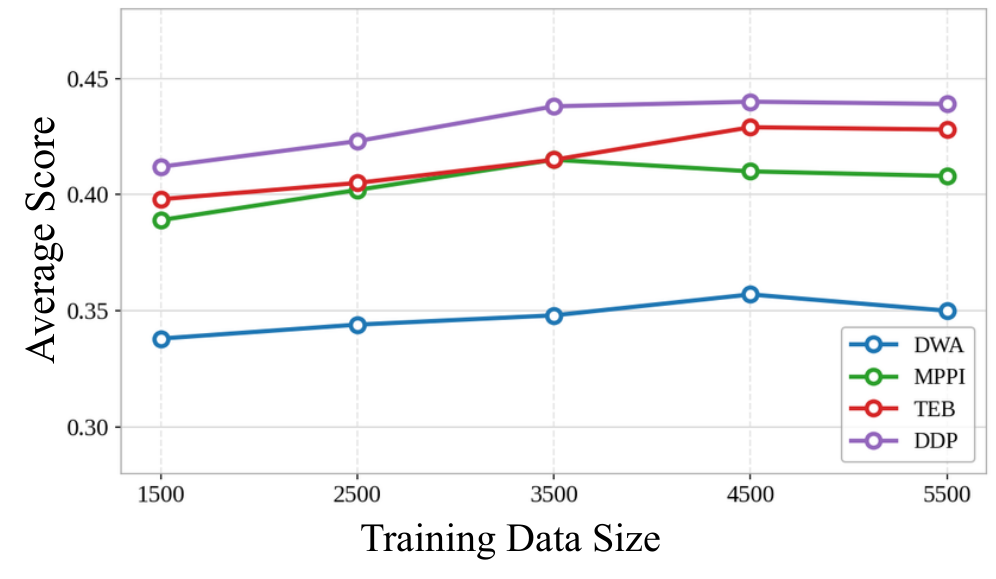}\\[-2mm]
  \caption{Effect of Training Data Size (\textsc{applv-sl}).}
  \vspace{-3mm}
  \label{fig:ablation}
\end{figure}

\subsection{Effect of Training Data Size}
We investigate the impact of training data size on \textsc{applv-sl} by varying samples from 1,500 to 5,500. Fig.~\ref{fig:ablation} shows that performance does not monotonically improve with data size. All planners reach peak performance at moderate data sizes and plateau or degrade thereafter. This suggests that additional data primarily helps the VLM learn feature representations rather than memorize examples. Once core features are captured, further data provides diminishing returns.

\section{Conclusion}
\label{sec:Conclusions}
We present \textsc{applv}, which leverages vision-language models for adaptive navigation planner parameter learning. This work demonstrates that pre-trained VLMs possess powerful visual reasoning capabilities for robot navigation. By predicting planner parameters rather than direct actions, \textsc{applv} enables dynamic adaptation while maintaining safety assurances of classical planners.

Experiments across four planners on the BARN benchmark and physical robots show that \textsc{applv} consistently outperforms existing methods, including \textsc{applr}, Heuristic Expert, and Transformer BC. This validates that vision-language models can effectively understand navigation scenarios and make informed parameter decisions, offering a promising direction for adaptive robot navigation that combines foundation model strengths with classical planning reliability.

\section*{acknowledgements}
This work has taken place in the RobotiXX Laboratory at George Mason University. RobotiXX research is supported by National Science Foundation (NSF, 2350352), Army Research Office (ARO, W911NF2320004, W911NF2520011), Army Ground Vehicle Systems Center (GVSC), Google DeepMind (GDM), Microsoft Research (MSR), Clearpath Robotics, FrodoBots Lab, Raytheon Technologies (RTX), Tangenta, 4-VA, Mason Innovation Exchange (MIX), and Walmart.

\IEEEpeerreviewmaketitle

\bibliographystyle{IEEEtran}
\bibliography{ref.bib}

\end{document}